\documentclass[review]{elsarticle}

\usepackage{lineno,hyperref}
\usepackage{graphicx}
\usepackage{subfigure}
\usepackage{multirow}
\usepackage{amsmath}
\usepackage{amssymb}
\usepackage{tabularx}
\usepackage{CJKutf8}
\begin{document}

\begin{frontmatter}

\title{SASICM: A Multi-Task Benchmark For Subtext Recognition}
% \tnotetext[mytitlenote]{Fully documented templates are available in the elsarticle package on \href{http://www.ctan.org/tex-archive/macros/latex/contrib/elsarticle}{CTAN}.}

% \author{Hua Yan\fnref{myfootnote}}
% \address{}
% \fntext[myfootnote]{Since 1880.}

\author[mymainaddress,mysecondmainaddress]{Hua Yan}
\ead{mg1937028@smail.nju.edu.cn}
% \ead[url]{https://github.com/William-YanHua}
\author[mymainaddress]{Feng Han}
\ead{fenghan@small.nju.edu.cn}
\author[mymainaddress]{Junyi An}
\ead{dz1833001@smail.nju.edu.cn}
\author[mymainaddress,mysecondmainaddress]{Weikang Xiao}
\ead{xiaowk5516@163.com}
\author[mymainaddress,mythirdaryaddress]{Jian Zhao}
\ead{jianzhao@nju.edu.cn}
\author[mymainaddress,mysecondmainaddress]{Furao Shen\corref{mycorrespondingauthor}}
\cortext[mycorrespondingauthor]{Corresponding author}
\ead{frshen@nju.edu.cn}

\address[mymainaddress]{State Key Laboratory for Novel Software Technology, Nanjing University, China}
\address[mysecondmainaddress]{College of Artificial Intelligence, Nanjing University, China}
\address[mythirdaryaddress]{School of Electronic Science and Engineering, Nanjing University, China}

\begin{abstract}
Subtext is a kind of deep semantics which can be acquired after one or more rounds of expression transformation. As a popular way of expressing one's intentions, it is well worth studying. In this paper, we try to make computers understand whether there is a subtext by means of machine learning. We build a Chinese dataset whose source data comes from the popular social media (e.g. Weibo, Netease Music, Zhihu, and Bilibili). In addition, we also build a baseline model called SASICM to deal with subtext recognition. The $F_1$ score of SASICM$_g$, whose pretrained model is GloVe, is as high as 64.37\%, which is 3.97\% higher than that of $BERT$ based model, 12.7\% higher than that of traditional methods on average, including support vector machine, logistic regression classifier, maximum entropy classifier, naive bayes classifier and decision tree and 2.39\% higher than that of the state-of-the-art, including MARIN and BTM. The $F_1$ score of SASICM$_{BERT}$, whose pretrained model is BERT, is 65.12\%, which is 0.75\% higher than that of SASICM$_g$. The accuracy rates of SASICM$_g$ and SASICM$_{BERT}$ are 71.16\% and 70.76\%, respectively, which can compete with those of other methods which are mentioned before.
\end{abstract}

\begin{keyword}
Subtext\sep deep semantics\sep expression transformation\sep SASICM\sep subtext recognition.
\MSC[2010] 00-01\sep  99-00
\end{keyword}

\end{frontmatter}

% \linenumbers

\begin{CJK*}{UTF8}{gbsn}
\section{Introduction}
\label{Introduction}
Subtext is a kind of deep semantics, which cannot be obtained from the text sequence directly. We put forward the concept of subtext analysis and we classify it into two phases, including subtext recognition and subtext recorvery. Furthermore, subtext recognition is a sub-task of text classification. Text classification is an important branch of natural language processing, which is greatly focused on sentiment analysis, metaphor recognition, etc.
% Done or a speech or dialogue->a speech or a dialogue
The definitions of subtext in Chinese and English\footnote{Definition in English: \url{https://en.wiktionary.org/wiki/subtext},\par \quad Definition in Chinese:  	\url{https://baike.baidu.com/item/\%E6\%BD\%9C\%E5\%8F\%B0\\\%E8\%AF\%8D/82560?fr=aladdin}} meet high-level agreement that can be summarized as ``implicit meaning of a text, often a literary one, a speech, or a dialogue''. \par
We describe subtext as a kind of deep semantics that can be obtained after one or more rounds of expression transformation. Formally speaking, expression A is a simple expression which directly describes the basic true intention, while expression B is an expression which indirectly conveys the intention. There is a transformation path from B to A by two kinds of transformation methods:
%, and there are several transformation methods: 
1. replace rhetorical words with their corresponding original words. For example, eliminate metaphorical words by replacing them with ontology, replace sarcastic words with negative forms and so on; 2. reason with the substitution expression after step 1, and draw a conclusion. For instance, replace the abstract meaning with the original meaning according to the context or background knowledge.\par
For example, in Chinese,  ``你的心里有一道墙，我要跨过这道墙~(If there is a wall in your heart, I want to cross it)'' which means that ``I'' will overcome difficulties and become your boyfriend/girlfriend, not that ``I'' want to cross a wall. 
The transformation process is as follows: firstly, replace the metaphorical words ``墙~(the wall)'' and ``cross~(cross)'' with ``difficulties'' and ``overcome'' respectively; secondly, reason according to the context. We get that the speaker want to overcome difficulties to get into another girl's heart. Finally, we come to the conclusion that this sentence means ``I want to be your girlfriend/boyfriend, I will overcome difficulties''. Besides, ``I burn you'' is another subtext in English. When we use it in a debate, it means that we win the debate. \par
This paper aims to be an enlightening research to judge whether a given sentence is subtext. We define this problem as \emph{subtext recognition}, which is a sub-field of text classification. The research of the transformation process for extracting subtext is defined as \emph{subtext analysis}. Obviously, subtext recognition is the precursor step of subtext analysis.  \par
% Done 介绍sentiment analysis
To judge whether a sentence is subtext, this paper constructs a Chinese corpus. After correlation analysis, we find that subtext is highly correlated with sarcasm and metaphor. Therefore, we decide to construct a multi-task model, similar to \citep{majumder2019sentiment, DBLP:journals/access/JinWMYM20, DBLP:journals/ijon/AkhtarGE20} that use multi-task frameworks to improve performance of classification.
\par
% Done 简要概括一下本文的核心点
\textbf{Contribution}: As far as we know, we are the first to analyze whether a sentence is subtext or not. Such analysis empowers machines to know what people really mean, which can make machine translation and sentiment analysis more accurate. Our contribution can be summarized as follows: 
\begin{enumerate}
	\item[-] We put forward text subtext analysis in the field of natural language processing.
	\item[-] We build a \textbf{C}hinese \textbf{s}ubtext \textbf{d}ataset (CSD-Dataset) from popular social media, including Weibo\footnote{\url{https://weibo.com/}}, Zhihu\footnote{\url{https://www.zhihu.com}}, Netease Music\footnote{\url{https://music.163.com/}}, and Bilibili\footnote{\url{https://www.bilibili.com/}}, and evaluate its quality. 
	\item[-] We propose a new reliability evaluation metrics TAE score to help judge whether a dataset is valid when it comes to an imbalanced label distribution in Two-Round Annotation.
	\item[-] We establish a multi-task benchmark SASICM for subtext recognition, which obtains a higher F$_1$ score than the other comparison models.
\end{enumerate}

\section{Related Work}
% 介绍一下基本的文本分类技术
% 相关的技术领域：文本分类：情感分析、反讽识别、比喻识别等。
\subsection{Sentiment Analysis}
Sentiment analysis aims to analyze people's opinions,
sentiments, evaluations, appraisals, attitudes and emotions towards entities such as products,
services, organizations, individuals, issues, events, topics and their  attributes \citep{DBLP:journals/widm/ZhangWL18, liu2012sentiment}. Sentiment analysis \citep{bataa2019investigation, schmitt2018joint, tang2019progressive} is regarded as a text classification task: positive (1), negative (-1) and nerual (0). Schmitt et al. \citep{schmitt2018joint} combined Convolutional Neural Networks (CNNs) and FastText to construct models and got the state of the art at that time. Tang et al. \citep{tang2019progressive}, Luo et al. \citep{luo2019towards} and Bao et al. \citep{bao2019attention} all use 
% are all use -> all use
Bi-directional Long Short Term Memory (Bi-LSTM) and attention Mechanism as parts of their basic model blocks, where Tang et al. \citep{tang2019progressive} introduced external supervised learning to improve the performance,  and Luo et al. \citep{luo2019towards} introduced sentiment embeddings and semantic embeddings to capture the sentiment feature better.
%Done better capture the sentiment feature ->capture the sentiment feature better
Bao et al.\cite{bao2019attention} modified the regularization of attention and greatly improved the system performance. Liang et al.\cite{liang2019context} introduced the context aware embedding to make it easier to capture the context semantics based on Bi-LSTM as well. 
\subsection{Sarcasm Detection}
Sarcasm detection \citep{DBLP:journals/corr/JoshiBC16, ghosh2020report, DBLP:conf/acl/MishraKNDB16} can be regarded as a subfield of sentiment analysis. The work \cite{zhang2016tweet} used GloVe as the embedding and RNNs as the basic block to construct a deep learning model. Tay et al.\cite{tay2018reasoning} proposed to combine the sequence context feature and intra-attention, and got the best performance at that time. 
%where sequence context feature was obtained by the LSTM model and intra-attention was computed directly from original embeddings.
Because of the sentence representations processed by LSTM are similar, instead of using the stack-like structure, Tay et al. \citep{tay2018reasoning} fed the pre-trained embeddings into attention-layer directly which is called intra-attention designed by themselves. In \citep{hazarika2018cascade}, the authors used CNNs to extract feature with max-pooling to enhenced this kind of feature. This paper assumes that the sentence representations processed by LSTM or GRU are similar, which is proved by our experiments.
\subsection{Metaphor Detection}
Metaphorical analysis focuses on exploring the relationship between two different concepts or fields  \citep{rei-etal-2017-grasping}. Take ``curing juvenile delinquency'' for example. We always view the crime (the target concept) in terms of the property of disease (the source concept). Metaphor detection \citep{jang-etal-2016-metaphor, bizzoni-lappin-2018-predicting} is an important part of metaphor analysis, and its main goal is to judge whether there is a metaphor in the text. The fine-grained task of metaphor detection \citep{stowe-palmer-2018-leveraging, mosolova-etal-2018-conditional}, which can also be called token-level metaphor detection, is to identify which part has metaphor. \par
Mao et al. \cite{mao2018word} used WordNet to extract the semantic feature and used cosine similarity as the computing technology to judge whether there was metaphor. Mao et al. \cite{mao2019end} pretrained a literal embedding, and then used Bi-LSTM and Attention as the main block to analyse the metaphor.
\subsection{Multi-Task Model}
Majumder at el. \cite{majumder2019sentiment} introduced a multi-task structure with Bi-directional Gated Recurrent Unit (Bi-GRU) and attention as the feature extractor, and tensor network as the multi-task confusion tool.
The performance in \citep{majumder2019sentiment} achieved the state of the art. Jin et al. \cite{DBLP:journals/access/JinWMYM20} combined the strength of CNNs, LSTMs and Multi-task structure. Akhtar et al. \cite{DBLP:journals/ijon/AkhtarGE20} executed term-extraction task and sentiment classification at the same time to improve the accuracy of sentiment classification.
\subsection{Concept Distinguish for Subtext Analysis}
In this subsection, we distinguish the differences among subtext analysis, sentiment analysis and metaphorical analysis in brief. \par
First of all, the goals of subtext analysis, sentiment analysis and metaphorical analysis are different. The purpose of subtext recognition is to judge whether there is another meaning besides what is conveyed by the original text. Sentiment analysis aims to judge the opinion or emotions, while metaphor analysis aims to judge the relationship between abstract entities and physical entities. Metaphors can be a way of representing subtexts, but not all metaphors are subtexts. Metaphor can be a subtext only if it can be derived a new meaning after reverting back to the original content. For example, if a person A said `you burned me' in a debate, it means that A was persuaded by someone (this is not subtext). In metaphor analysis, ``burning someone'' and ``winning someone'' have the same meaning. Moreover, this sentence cannot derive another meaning after replacing ``burning someone'' with ``winning someone''. In sentiment analysis, this sentence does not convey any emotions, nor does it have any opinion. Take ``被天使吻过的嗓音[大哭] (The voice is kissed by an angel [crying])'' for example. Metaphor analysis obtains that ``angel'' means ``nice thing or lucky''. The original sentence can be write as ``the voice is very nice [crying]'' after metaphor recovery, and it can be derived that ``this people sings so well that I am moved to tears'', which is the subtext. Moreover, the sentiment analysis just obtains that this sentence conveys a positive emotion `nice'. \par
Secondly, subtext analysis relies on context than metaphor analysis or sentiment analysis, which means it is harder than metaphor analysis and sentiment analysis. For instance, ``好家伙，起码九年。 (Wow! At least nine years!)'' has two kinds of explaination. When it follows a context that ``1个月一期，一共一百期，追个几年没问题了[doge]. (This program is shown once a month, a total of 100 episodes. I can watch for a few years [DOGE].)'', it means means that ``This program'' lasts a long time, and this commentor has a great patience. But when it follows a ``这发量，一看就是高级软件工程师。 (According to the hair quantity, he is a senior software engineer at a look.)'', it means that this person has a low number of taunts and is a programmer with long years. \par
Moreover, subtext analysis needs more backgrounds. For example, ``奶奶某一瞬间真像马王堆那位 (Grandma is really like Xin Zhui who lived in the Western Han dynasty of ancient China and was excavated in the tomb of Mawangdui.)'' can hard be analyzed if we do not know how the appearance of ``Xin Zhui'' when she was excavated. \par
    In conclusion, sentiment analysis just analyze the emotions conveyed by the commenter. Metaphor analysis only analyze the noumena and metaphor. Only when the sentence can derive a new meaning after metaphor recovery or sarcasm recorvery, it is a subtext. Subtext analysis aims to find whether there is another meaning. metaphor, sarcasm and other figurative methods are part of the expression ways of subtext. Moreover, subtext needs more context to help dinstinguish the other meaning. In addition, subtext often needs background knowledge to jugde the original meaning.

% 例子二补充：奶奶某一瞬间真像马王堆那位

\begin{table*}[!t]
    \centering
    \caption{\label{annotation-samples} Annotation samples, where ``cont'', ``subt'', ``sarc'', ``meta'', ``exag'', ``homo'', ``emot'', ``atti'', ``other'' stand for ``content'', ``subtext'', ``sarcasm'', ``metaphor'', ``exaggeration'', ``homophonic'', ``emotion'', ``attitudes'' and other ``kinds of rhetoric'' respectively.}
    \resizebox{\textwidth}{!}{
        \begin{tabular}{p{0.4\textwidth}ccccccccc}
            \hline
            \textbf{comment} & \textbf{cont}  & \textbf{subt} & \textbf{sarc} & \textbf{meta}&\textbf{exag} &\textbf{homo} &\textbf{emot} & \textbf{atti} & \textbf{other}\\
            \hline
            {暗恋？一个人的兵荒马乱.~(Secret love? It is a man of war.)} & \multirow{2}*{null} & \multirow{2}*{1} & \multirow{2}*{-1} & \multirow{2}*{1} & \multirow{2}*{0} & \multirow{2}*{-1} & \multirow{2}*{None} & \multirow{2}*{0} & \multirow{2}*{-1}\\
            {你隔岸观火 却不救我~(You watched the fire from the other side but didn’t save me)} & \multirow{3}*{null} & \multirow{3}*{1} & \multirow{3}*{0} & \multirow{3}*{1} & \multirow{3}*{-1} & \multirow{3}*{-1} & \multirow{3}*{sad} & \multirow{3}*{-1} & \multirow{3}*{-1}\\
            \hline
        \end{tabular}
    }
    
\end{table*}
\section{CSD-Dataset}
% 对数据的来源做一个介绍
After exploring a large number of websites, we find that it is more likely to include subtext in the comment data. Therefore, we have chosen some popular social media to collect source data, such as Weibo, Zhihu, Netease Music and Bilibili. 
%Done 名称Bilibili和前面不一致

\subsection{Data Collection}
% 对数据的处理做一个介绍
We grab the comment data from the hot lists of the four major websites, which arouse people's attention. To use it in multi-turns of communication analysis, we retain the structure information of the source comment. This information includes annotation, annotation ID, parent ID and parent content, wherein the parent content is only referred to as content hereinafter. In the end, we collected about 70,000 comments. Moreover, our dataset has been anonymized so that it does not involve the user's personal identity information.
\subsection{Annotation}

To prevent subjective influence, each comment was labeled by three people independently and then the three different labels were checked by other people. Some of annotation samples are displayed in Table \ref{annotation-samples}.

We annotate a comment with seven kinds of information: \emph{sarcasm}, \emph{metaphor}, \emph{exaggeration}, \emph{homophonic}, \emph{attitude}, \emph{emotion} and \emph{other} information, among which \emph{other} is the unified category of other rhetorical methods, with a number less than 50. Subtext, sarcasm and metaphor are marked with three tags: Tag (1) means that the sentence contains subtext/sarcasm/metaphor; Tag (-1) means that the sentence does not contain subtext/sarcasm/metaphor; Tag (0) means that unsure, similar to~ \citep{mishra2017learning, DBLP:conf/rocling/LinH16}. In addition, we also annotate the original meaning of subtext, the objects of metaphor, the noumenon of metaphor and satirical words. To the best of our knowledge, this is the first work to label exaggerated and harmonic information. According to the previous work for text classification, this paper marks the classes -1 (not included), 0 (uncertain) and 1 (definitely included).  We label 8 kinds of emotional information: \textsl{anger, fear, disgust, trust, joy, surprise, anticipation, sad} as ~\citep{kant2018practical, plutchik1984emotions} did. In addition, we also add \textsl{None} to indicate that there is no obvious emotional expression in text. The attitudes are annotated like~\citep{nakov2019semeval2013, rosenthal-etal-2017-semeval}.
Eventually, we get 8,843 annoated comments after removing useless data.

\begin{table*}
    \centering
    
    \caption{\label{annotation-score} The score of different evaluation methods.}
    \resizebox{\textwidth}{!}{
    \begin{tabular}{lcccccc}
        \hline
        \textbf{type} & \textbf{sarcasm}  & \textbf{metaphor} & \textbf{subtext} & \textbf{exaggeration} & \textbf{homophonic} & \textbf{other}\\
        \hline
        \textbf{Kappa} & 0.60 & 0.60 & 0.60 & 0.71 & 0.61 & 0.26\\
        \textbf{TAE} & 0.81 & 0.56 & 0.50 & 0.88 & 0.95 & 0.93\\
        \hline
    \end{tabular}
    }
\end{table*}

\subsection{Quality Evaluation}
% 对标注后的数据进行统计分析：就是统计在不同网站中subtext出现的概率、sarcasm出现的概率、metaphor出现的概率，以及sarcasm、metaphor和subtext之间出现的相关性
To ensure the annotation quality, this paper adopts Two-Stages labeling to avoid subjectivity and uses two methods to evaluate the reliability. The Two-Stages labeling is divided into two phases: In the first phase, three people are asked to label respectively and independently;
%Done independent->independently
In the second phase, the fourth person annotates the text according to the three labeling results. There are several situations in the second phase: 1. all the labels are the same, then we adopt them;
2. all the labels are different, then we delete this comment or relabel this comment; 3. part of labels are different, then we re-annotate it.  \par
\begin{figure}
	\centering
	\begin{minipage}[t]{\textwidth}
		\resizebox{\textwidth}{!}{
			\subfigure[Kappa Changes with Agreements.]{
				\includegraphics[width=0.5\textwidth]{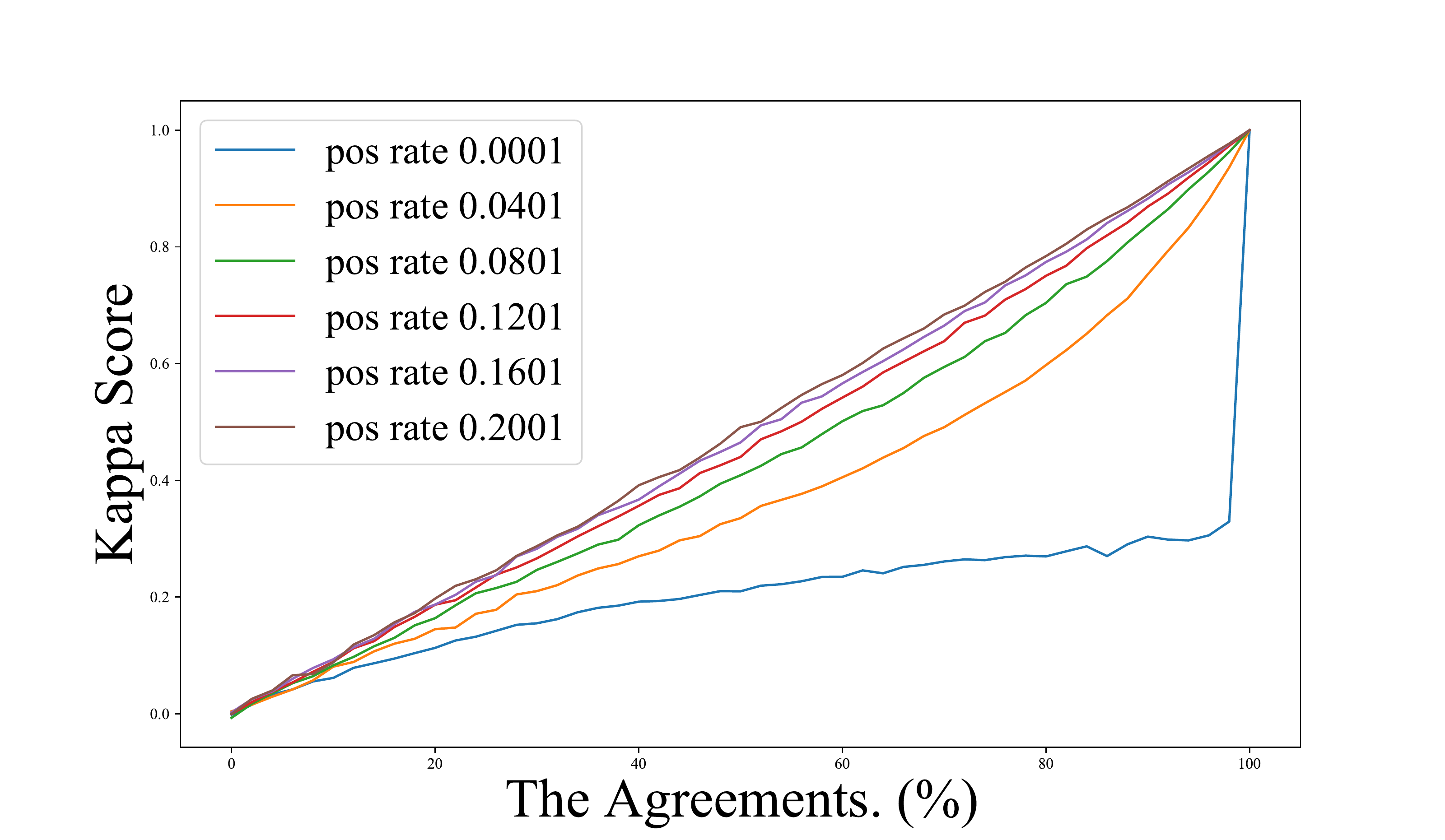}
				\label{Fig: Kappas}
			} ~~~
			\subfigure[Accuracy Changes with Agreements.]{
				\includegraphics[width=0.5\textwidth]{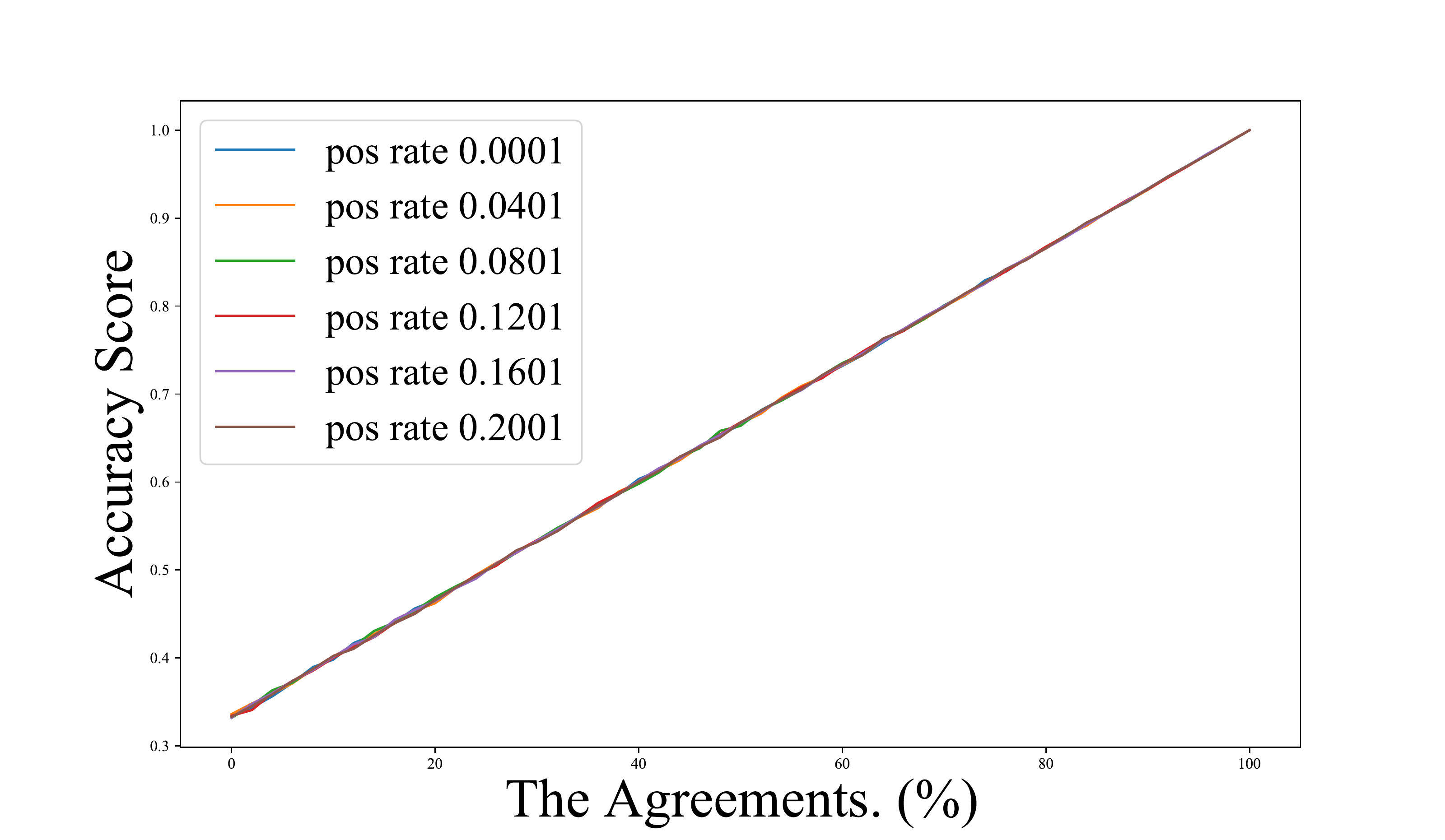}
				\label{Fig: Accuracy}
			}
		}
		\resizebox{\textwidth}{!}{
			\subfigure[TAE Changes with Agreements.] {
				\includegraphics[width=0.5\textwidth]{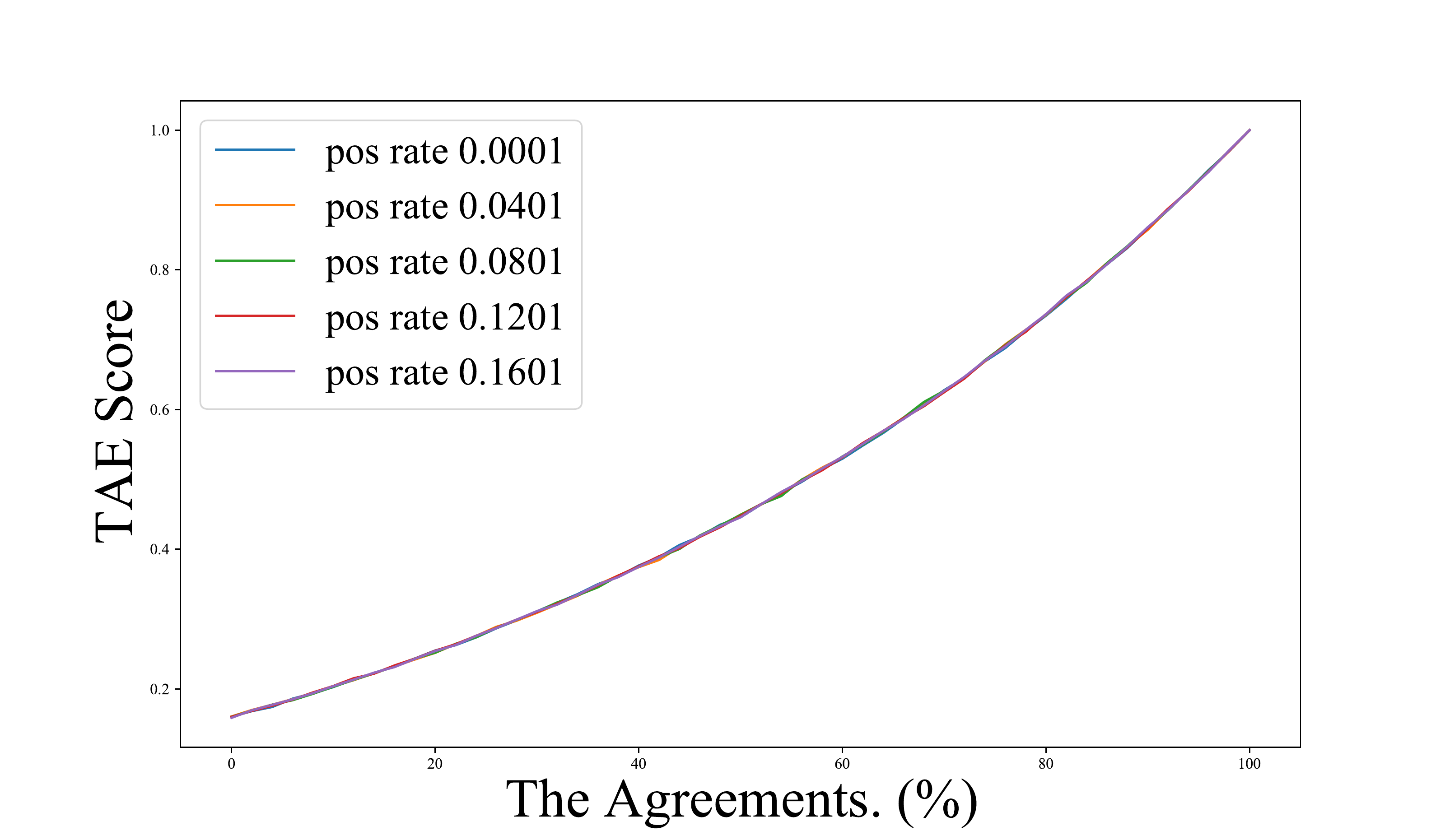}
				\label{Fig: TAE}
			} ~~~
			\subfigure[Comparison Of Kappa, Accuracy and TAE.] {
				\includegraphics[width=0.5\textwidth]{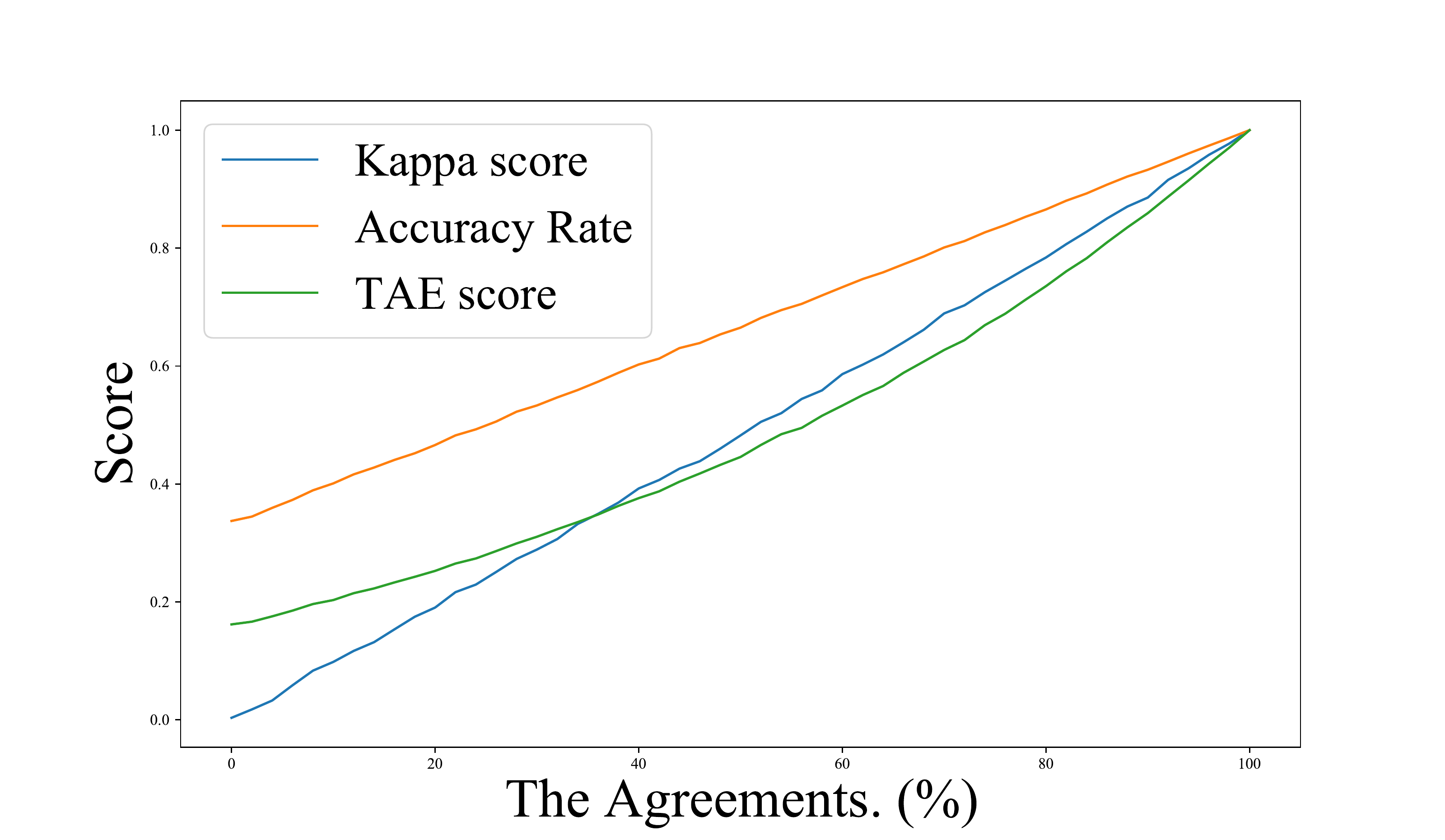}
				\label{Fig: Comparison}
			}
		}
	\end{minipage}
	\caption{Validation Metrics Comparison. Pos rate means the ratio of positive samples and all samples.}
	\label{Fig: TAE-Comparison}
\end{figure}

%Done 前面是he后面是himself和herself，不一致
To evaluate the reliability, this paper uses Kappa score as ~\citep{ghanem2019idat, khodak2017large, webster2018mind} did. However, Kappa score is usually used under completely independent conditions, which is not suitable for our two rounds of labeling treatment. What is worse, there is a problem with Kappa score ~\citep{artstein2008inter, sim2005kappa}: if the data is extremely imbalanced, the Kappa score will be low, even the annotators meet high agreement. Therefore, this paper introduces an evaluation metric which is termed the \textbf{T}wo-Rounds \textbf{A}nnotation \textbf{E}valuation (TAE). \\
\textbf{TAE Score}: To compute TAE score, we first define the two values for single annotation records.
\begin{enumerate}
	\item[-] Agreement: The ratio of the labeling results of the first round and those of the second round to the total number of labeling. Let $L_1 = \{l_{11}, l_{12},\cdots,l_{1n}\}$ be the labeling results of the first round annotation, $l_2$ be the labeling result of the second round annotation. The agreement for the $i$-th records is computed as:
	\begin{align}
		agr_i = \frac{|\{l_1j| l_{1j} == l_2; j = 1,2,\cdots, n\}|}{n}.
	\end{align}
	\item[-] Randomness: The ratio of the label types between the labeling results of the first round and the labeling results of the second round to the total label types. Let $\rm{ls}(s)$ be the function of turnning a list into a set, and $s$ be a list. Let $\rm{Li}_1 = [l_{11},l_{12}, \cdots, l_{1n}]$ be the tabular form of $L_1$, and $\rm{Li}_2$ to be tabular form of $l_2$. The randomness for the $i$-th records is computed as:
	\begin{align}
	rad_i = \frac{\rm{ls}(\rm{Li}_1)\setminus\rm{ls}(\rm{Li}_2)}{\rm{ls}(\rm{Li}_1)\cup \rm{ls}(\rm{Li}_2)}
	\end{align}
\end{enumerate}
To be a validation metric, TAE should satisfy the following properties:
\begin{enumerate}
	\item[-] Monotony. TAE score should be monotonically increasing with respect to the agreement. TAE score should be monotonically decreasing with respect to the randomness as well.
	\item[-] Boundness. TAE score needs to be a bounded function about randomness and consistency, so that we can measure whether a data set is reliable.
	\item[-] Independence. TAE score should be independent from the ratio of positive samples and negtive samples, which is the main shortcomings of Kappa score.
\end{enumerate}
Consequently, we define TAE score as follows:
\begin{align}
	\rm{TAE} &= \frac{\exp(agr-rad)-1/e}{e-1/e} \\
	agr &= \frac{\sum_{i=1}^n agr_i}{n} \\
	rad &= \frac{\sum_{i=1}^n rad_i}{n}.
\end{align}
To illustrate TAE score is valid, we make some simulation experiments. We execute the simulations under different ratios of positive samples and negtive samples. Each simulation experiment describes how the validation score changes with respect to the agreements in three classifications. The results is shown in Figure \ref{Fig: TAE-Comparison}.
Figure \ref{Fig: Kappas}, \ref{Fig: Accuracy} and \ref{Fig: TAE} show that the curve changes of Kappa score, accuracy rate and TAE score with respect to the agreement under different ratios of positive samples and negative samples, respectively. Figure \ref{Fig: Comparison} shows that the performances of accuracy rate, Kappa score and TAE score in the same balanced ratio of postive samples and negative samples, and the ratio is 0.2. Figure \ref{Fig: Accuracy} shows that accuracy is linear increasing with respect to agreement. It does not consider the influence of randomness. Figure \ref{Fig: Kappas} and Figure \ref{Fig: TAE} show that Kappa score and TAE score is non-linear increasing with respect to the agreement. Both of them consider the influence of randomness. Moreover, from Figure \ref{Fig: Kappas}, we can find that Kappa score will be low when agreement is less than a high threshold about 96\% under the setting of an extremely imbalanced label distribution, then it will explore. However, the TAE score will not be influenced by the level of imbalance, just like accuracy rate. Figure \ref{Fig: Comparison} shows that accuracy rate is higher than Kappa score and TAE score under the situation of the labels distributing balanced. Furthermore, TAE has a similar performance as Kappa. Therefore, TAE can be used to evaluate the validation of our dataset. We point out that the Kappa score means high reliability if it is above 0.6. The corresponding score of TAE is 0.53 with the same agreement and randomness as Kappa whose score is 0.6.

\subsection{Analysis}
\begin{figure}[h]
	\centering
	\includegraphics[width=0.7\linewidth]{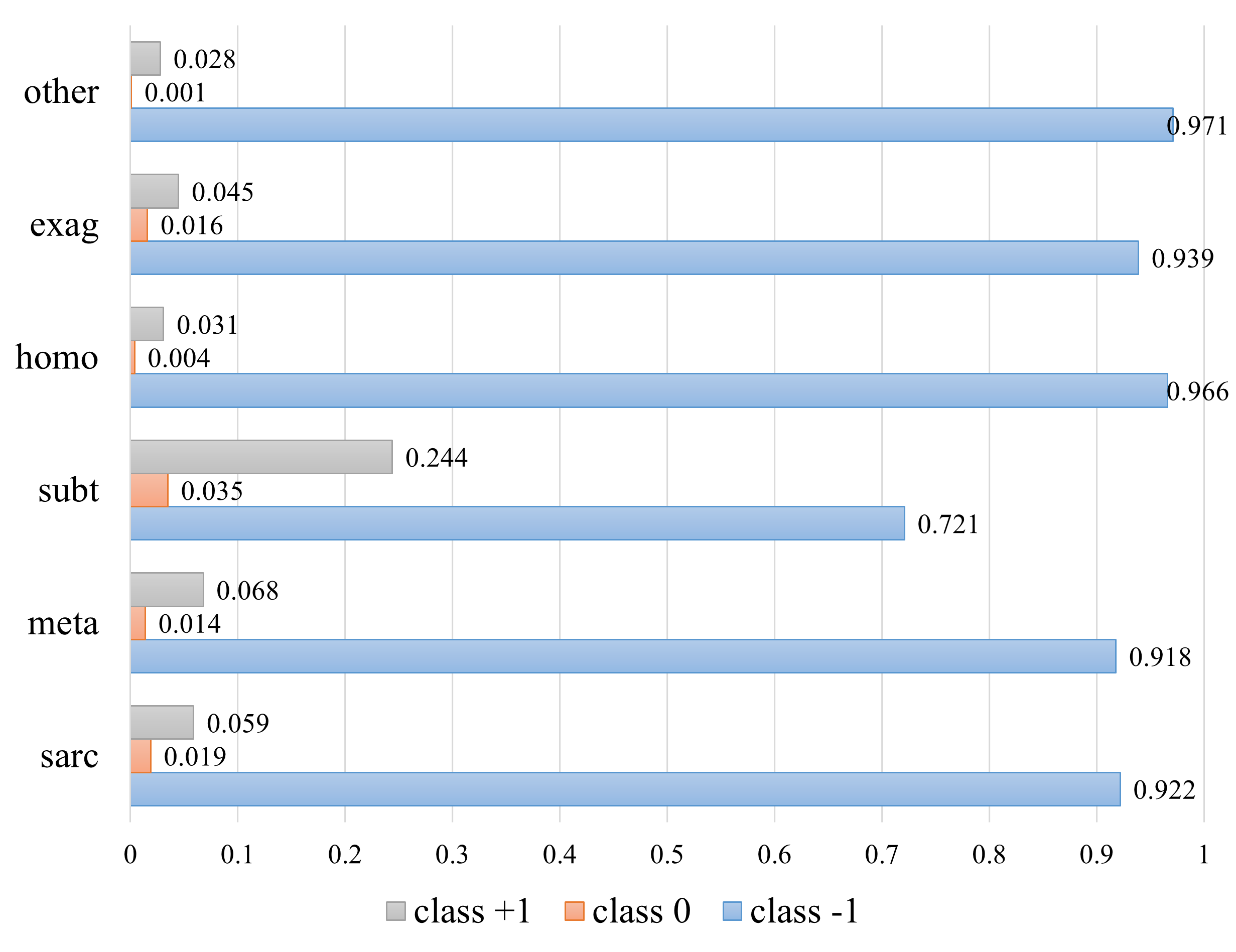}
	\caption{The ratio of different labeling information. The horizontal axis represents ratio value, and the vertical axis represents different annotation information.}
	\label{graph:classes number}
	\includegraphics[width=0.7\textwidth]{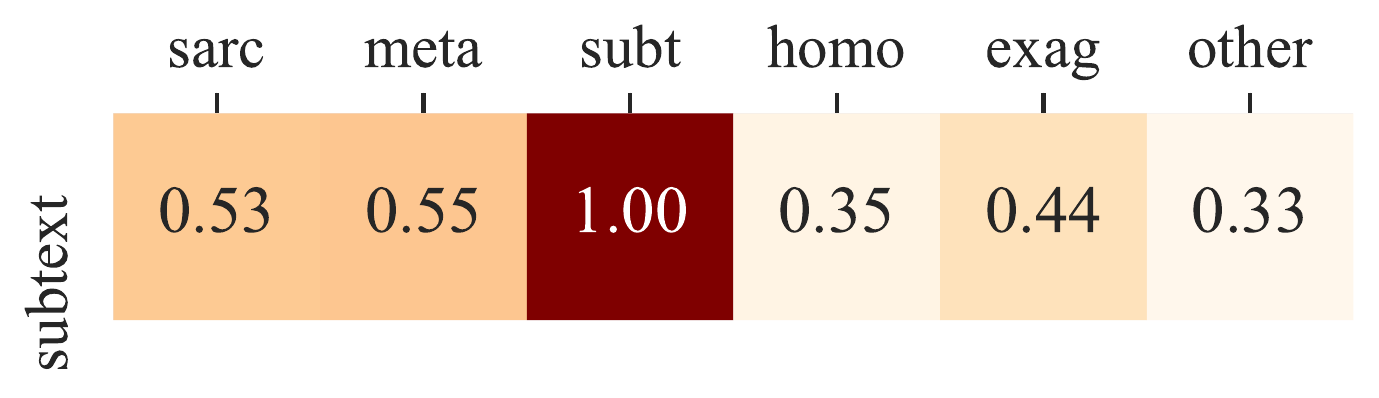}
	\caption{The correspondence coefficient of different label.}
	\label{graph: corr-different label}
\end{figure}

The Kappa score and TAE score of the results of annotated by us are shown in Table~\ref{annotation-score}. 
Both of them convey the quality of our dataset.
The score of $other$ is high in TAE, but is lower in Kappa. Figure \ref{graph:classes number}
displays the ratio of different classes for different labeling information. It is obvious that the distribution of different classes is extremely imbalanced, which will cause a problem of high agreement but low score in Kappa as declared in~\citep{sim2005kappa, feinstein1990high}. 
Therefore, we combine the evaluation result of TAE score and Kappa score, and come to the conclusion that our corpus is reliable. The scores of TAE and Kappa of subtext show that it is more tough to annotate than other information. In addition, we make a correlation analysis on different classes, and the correlation coefficients are shown in Figure \ref{graph: corr-different label}. Subtext is highly related to sarcasm (0.53) and metaphor (0.55).  
%     The coefficients between subtext and sarcasm is 0.53, and the coefficients between subtext and metaphor is 0.55. 
Considering the correlation, we set up a multi-task framework in the experiment.

\section{SASICM: A Multi-Task Model}
\begin{figure*}[!t]
    \centering
    \includegraphics[width=\linewidth]{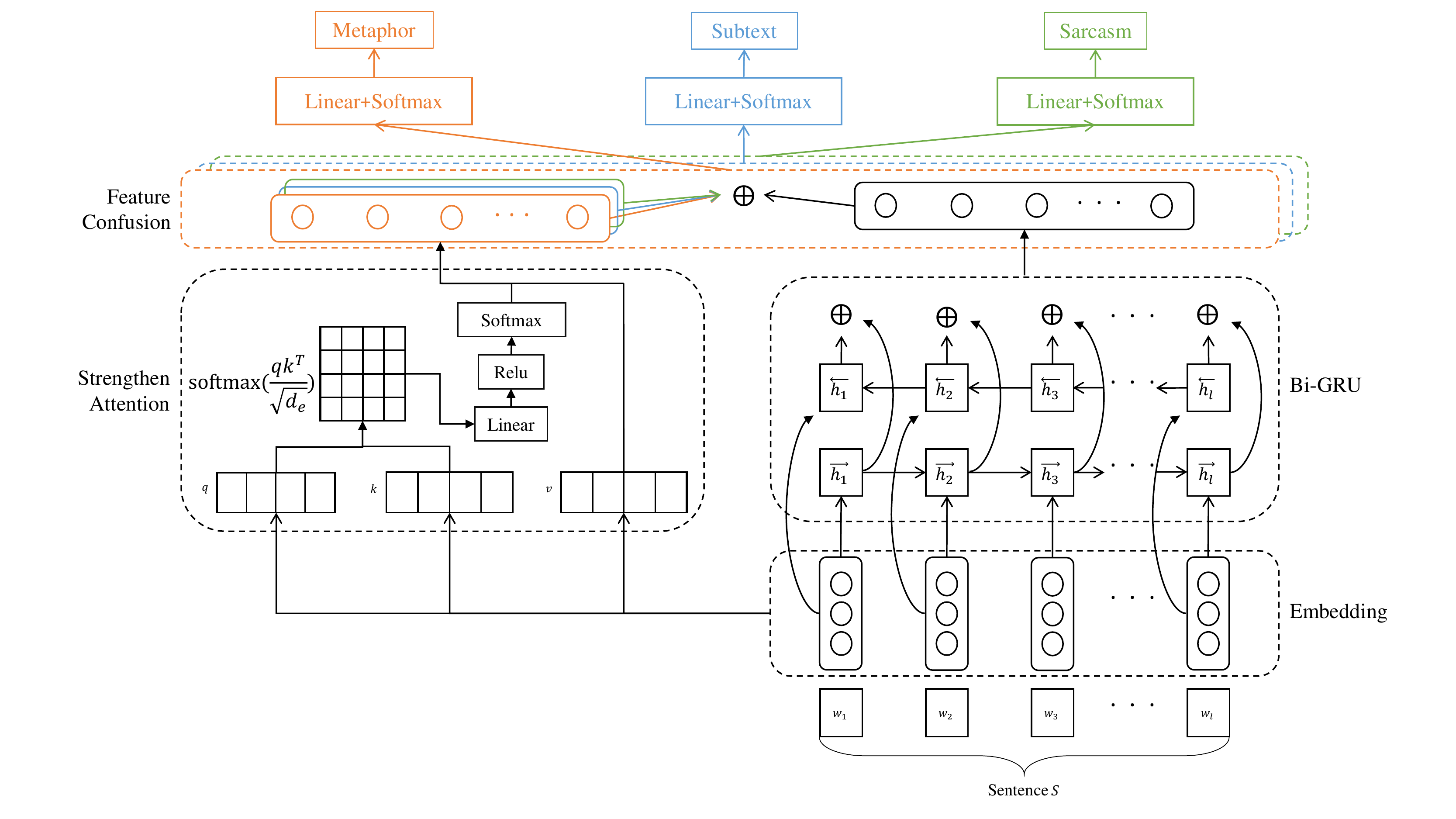}
    \caption{The structure of our model.}
    \label{graph:model-structure}
\end{figure*}
\subsection{Overview}
In order to deal with subtext recognition, we constructed a multi-task baseline to better understand this problem. The main idea is drawn from \citep{tay2018reasoning, majumder2019sentiment, tang2019progressive}. Our model contains one embedding layer, one strengthen attention layer, one bi-directional GRU layer, three feature confusion layers and three prediction layers.
The model structure is shown in Figure \ref{graph:model-structure}. We call our model as \textbf{S}trenghten \textbf{A}ttention based \textbf{S}equence and \textbf{I}ntra-Attention \textbf{C}onfused \textbf{M}ulti-Task Model (SASICM). We  will describe the details of each component in following subsections. 

\subsection{Task Formalization}
Let $\mathit{S}=\{w_1, w_2,\cdots, w_l\}$ be the input text, where $w_i$ is the $i$-th word within the text of vocabulary $\mathit{V}$, and the number of words in the input text is $l$. The goal of subtext recognition is to predict a label $y~(y\in\mathit{R}^3)$, of which each dimension denotes it is \textit{not subtext}, \textit{is not sure} and \textit{is subtext}, respectively. In order to get more information and features, we add two tasks: sarcasm detection and metaphor detection, as the correlation between these two tasks and subtext shown in Figure  \ref{graph: corr-different label} is higher than 0.5.
The outputs of sarcasm detection and metaphor detection are similar to subtext, which are denoted as $\hat{y}_{sarc}$ and $\hat{y}_{meta}$ where $\hat{y}_{sarc}\in \{1,0,-1\}$ and $\hat{y}_{meta}\in \{1,0,-1\}$, respectively.

\subsection{Embedding Layer}
We use Glove \citep{pennington2014glove} and BERT \citep{DBLP:conf/naacl/DevlinCLT19} as our embedded layer model for pre-training, and then fine-tune it during training. The embedding size $d_{e}$ of Glove is set to be 300 empirically. The embedding size of BERT is set to be 768 according to its original paper.
\subsection{Strengthen Attention}
Strengthen Attention is used to maximize the difference between important words and unimportant words, and to expand the difference of attentions. Based on self-attention, we use softmax function to obtain attention $att$ from similarity $s_i$. Then, we apply linear projection to expand $att$ then obtain $\it{scaleds_i}$. Finally, we feed the $\it{scaleds_i}$ into a relu function to filter the unimportances word to abtain a truly important attention $att_t$. To ensure the total probability is 1, we rescaled $att_t$ into $[0,1]$ by softmax function. Formula (\ref{strength-attention:1}) to (\ref{strength-attention:2}) show how to compute strengthen attention.
\begin{align}
    s_{i,j} & = \frac{w_iQ\cdot (w_jK)^T}{\sqrt{d_h}} \label{strength-attention:1}\\
    s_i & = \left[s_{i,0}, s_{i, 1}, \cdots, s_{i, l-1}\right]\\
    \it{scaleds}_i & = t_t \cdot  \left( \mathrm{softmax}\left(s_i\right)-c \right) \\
    att_i &= \mathrm{softmax}\left(\mathrm{relu}\left(\it{scaleds}_i\right)\right) \\
    r_i &= att_i\cdot w_jV \label{strength-attention:2}
\end{align}
where $Q,K,V \in \mathit{R}^{d_e\times d_h}$ are projection parameters of query, key and value in respective. $d_e$ is the embedding size and $d_h$ is the hidden size. $\mathrm{relu}(x) = x$ if $x > 0$, else $\mathrm{relu}(x)=0$.
$c \in \mathit{R^{l\times d_h}}$ and $t_t \in \mathit{R^{l \times d_h}}$ are real value parameters. Then $r_i$ is the input into a dense layer to obtain a sentence representation with attention $r_{fa}$ as
\begin{align}
r_{fa} = \mathrm{softmax}(r_i\cdot \textbf{1}) \label{fea-rep-with-attention}
\end{align}
where $\textbf{1}$ is a column vector with 1 in all dimension.
\subsection{Bi-directional GRU Model}
In this model, we use the original GRU model to get context feature, and just use the final hidden states as the uni-directional feature. We denote the final hidden states of the forward directional and the backward directional as $h_{ff}$ and $h_{fb}$. In the output of this layer, we concatenate them as $h_c = [h_{ff}, h_{fb}]$. 
The uni-process is as follows:
\begin{align}
& r_t = \sigma(W_r\cdot[h_{t-1}, x_t])\\
& z_t = \sigma\left(W_z\cdot[h_{t-1}, x_t]\right) \\
& \tilde{h}_t = \tanh(W_{\tilde{h}}\cdot[r_{t}*h_{t-1}, x_t]) \\
& h_t = (1-z_t)* h_{t-1}+z_{t} * \tilde{h}_t
\end{align}
where $W_r \in \mathit{R}^{(d_h+d_{e}) \times d_h}$ , $W_z \in \mathit{R}^{(d_h+d_{e}) \times d_h}$, $W_{\tilde{h}} \in \mathit{R}^{(d_h + d_{e}) \times d_h}$ are weight matrices. $x_t \in \mathit{R}^{d_e}$ is the input at time $t$. $h_{t} \in \mathit{R}^{d_h}$ is the last hidden state of time $t$. $\sigma(x)$ is the sigmoid function
that $\sigma(x) = \frac{1}{1 + \exp(-x)}$. $d_h$ is the size of hidden states.
\subsection{Feature Confusion Layer}
After capturing the feature representation with attention $r_{fa}$ and the feature representation with context $h_c$, we concatenate them. In order to capture the deep fused feature, the concatenated representation is linearly mapped into the original space as (\ref{feature-confusion}) does.
\begin{align}
r_{\mathrm{taski}} = W_{fc|\mathrm{taski}} \cdot [h_c, r_{fa}] \label{feature-confusion}
\end{align}
where $W_{fc|\mathrm{taski}} \in \mathit{R}^{(2d_h + d_e) \times (2d_h + d_e)}$ is the weight matrix for feature confusion. $r_{\mathrm{taski}} \in \mathit{R}^{2d_h + d_e}$ stands for the outputs of subtext recognition, metaphor detection or sarcasm detection.
\subsection{Prediction Layer And Loss Function}
Finally, we use softmax to make a prediction as (\ref{prediction}) does. We treat the averaged cross-entropy as the loss function like (\ref{loss}). Besides, we add constraints to strengthen attention, $t_t$ should widen the gap of each attentions, and $c$ should be limited in $[\epsilon,1)$, where $\epsilon$ is a small number. Therefore, the final loss is as formula (\ref{loss}).
\begin{align}
\hat{y}_{taski} = &\mathrm{softmax}\left(W_p\cdot r_{taski} + b_p\right) \label{prediction}
\end{align}
\begin{align}
\mathfrak{L}(s, y) = &\frac{-1}{|\mathfrak{T}|N}\sum_{taski}\sum_{i=1}^{N}y_{taski}^{(i)}\log(\hat{y}_{taski}^{(i)}) \nonumber\\
& + \mathrm{relu}(c-1) + \mathrm{relu}(\epsilon-c) \nonumber \\
& + \alpha\cdot \mathrm{relu}(w_{thr}-t_t) \label{loss}
\end{align}
where $|\mathfrak{T}| \in \mathit{R}$ is the number of tasks. $y_{\mathrm{taski}}^{(i)} \in \mathit{R}^3$ is the ground truth label of the $i$th instance. $N$ is the number of instances. $\alpha \in \mathit{R}$ and $w_{thr} \in \mathit{R}$ are hyperparameters. 
\section{Experiment and Analysis} \label{Experiment and Analysis}
In this section, we mainly explore the problems of using SASICM: \\
\textbf{(1) Uni-task Framework Subtext Recognition:} We evaluate the performance of uni-task framework dealing with subtext recognition task only. The aim is to make a comparison of multi-tasks frame work.\\
\textbf{(2) Multi-task Framework Subtext Recognition:} We evaluate the performance of multi-task learning methods to comparing with other methods. In detail, we construct bi-task framework and tri-task framework, respectively. The aim is to validate the advantages of multi-task learning methods and set up multi-task baselines for CSD-Dataset.\\
\textbf{(3) Different Embedding Methods For Subtext Recognition:} We use GloVe \citep{pennington2014glove} (marked as SASICM$_g$) and BERT \citep{DBLP:conf/naacl/DevlinCLT19} (marked as SASICM$_{BERT}$) as our pre-train models, respectively. The aim is to evaluate the performance of different kinds of embedding. \\
\textbf{(4) Human Performance:} We take the same evaluation metrics to evaluate the performance of human on the labeling datasets. The aim is to see the upper limit and  to see whether our framework is work.
\subsection{Baselines}
In this section, we briefly review our baselines usedin the following experiments.\\
\textbf{Bag-Of-Word + Traditional Classifier} We use Bag-Of-Word as the input feature, and use traditional classifiers as the discriminators, including Support Vector Machine (SVM), Maximum Entropy Classifier (MEC), Naive Bayes (NB) and Logistic Regression (LR).\\
\textbf{BTM-Based Model} BTM is a framework for bi-task classification described in \citep{majumder2019sentiment}. BTM trains both sentiment analysis and sarcasm analysis tasks, and uses the results of sarcasm analysis as additional features to assist in identifying sentence sentiment. In our experiments, we train BTM in subtext recognition and sarcasm recognition (called BTMSS) and we train BTM in subtext recognition and metaphor recognition (called BTMSM). Moreover, we extend BTM to tri-task model by adding a metaphor task or sarcasm task (called BTM3). In addition, we also decrease the BTM into uni-task model for only recognizing the subtext (called BTMSubt).\\
%Done works->work
\textbf{MIARN-Based Model} MIARN is a uni-task framework for sarcasm detection introduced in \citep{tay2018reasoning}. MIARN feeds the word embedding into Intra-Attention and LSTMs to obtain inner feature and sequential feature respectively, then MIARN concatenates inner feature and sequential feature for classification. Like BTM, we extend MIARN into bi-task model (MIARNSS and MIARNSM) and tri-task model (MIARN3).\\
\textbf{BERT+Fully Connected Layer} BERT \citep{DBLP:conf/naacl/DevlinCLT19} is a widely used pre-train model which obtains the outstanding performance in many text classification tasks. We fine-tune BERT followed by a fully connected model according the original paper.\\
\textbf{GBP} To show that all the models have learned useful information, we add a probability-based random guess model, which is called \textbf{G}uess \textbf{B}y \textbf{P}robability (GBP), as one of comparison models. \\
\par
\subsection{Experimental Details}
In this section,  we introduce our experimental settings in detail, including dataset splits, hyper-parameters selection, and our evaluation metrics. \\
\textbf{Dataset splits} The ratio of the training set size to the test set size is 8:2. We train each model with the validation rate of 0.2 of the training set. Moreover, we split the data set according the label probability. To prevent our model from remembering the data order, we shuffle the training set before training. \\
\textbf{Hyper-parameters Selection} Due to the different sequence lengths in different data, it is necessary that fixing sequence length for the specific modality. Empirically, we counted the length of each piece of data, and used 99 percent of the previous length as the fixing length. During the experiments, we use the Nadam as the optimizer, and the learning rate is set to $10^{-3}$, the random seed is fixed to $10^5$, the batch size is 32, $\epsilon$ is $3\times10^{-3}$, $\alpha$ is $1\times10^{-2}$ and $w_{thr}$ is $5$. We use early stopping and dropout to reduce over-fitting, and the dropout rate is 0.1. We run all the models on one GPU (GeForce GTX 1080 Ti). \\
\textbf{Evaluation Metrics} Empirically, we adopted weighted $F_1$ score and accuracy as our evaluation metrics. To avoid randomness, we run all the models with 5-fold cross-validation, and repeat it 5 times.
\subsection{Result and Discussion}
\begin{table*}[h]
    \caption{\label{table: triple-task baseline} Baseline Results of Tri-Task. The result of f1 score (F1) and accuracy (acc), where ``p'' is precision and ``r'' is recall. The result marked by underline is the result of baseline model (SASICM). *$_m$ and *$_s$ are the results of metaphor and sarcasm respectively.}
    \centering
    \resizebox{\textwidth}{!}{
        \begin{tabular}{ccccccccccccc}
            \hline
            \multirow{2}{*}{\textbf{Model}} & \multicolumn{4}{c}{Subtext Task} & \multicolumn{4}{c}{Metaphor Task} & \multicolumn{4}{c}{Sarcasm Task}  \\
            \cline{2-13}
            & \textbf{p(\%)}  & \textbf{r(\%)} & \textbf{F$_1$(\%)} & \textbf{acc(\%)} & \textbf{p$_m$(\%)} &\textbf{r$_m$(\%)} & \textbf{F$_{1_m}$(\%)} & \textbf{acc$_m$(\%)}&\textbf{p$_s$(\%)}  & \textbf{r$_s$(\%)} & \textbf{F$_{1_s}$(\%)} & \textbf{acc$_s$(\%)}  \\
            \hline
            \fontsize{8pt}{5}{\textbf{\underline{SASICM$_{g}$}}} & 63.74 & 71.16 & \underline{64.37$^*$} & \underline{71.16$^*$} & 85.44 & 91.55 & \underline{\textbf{88.07}$^*$} & \underline{91.55$^*$} & \textbf{88.40} & 91.75 & \underline{88.75$^*$} & \underline{91.75$^*$} \\
            \fontsize{8pt}{5}{\textbf{\underline{SASICM$_{BERT}$}}} & \textbf{64.56} & 70.76 & \underline{\textbf{65.12}$^*$} & \underline{70.76$^*$} & \textbf{86.07} & 91.39 & \underline{\textbf{88.07}$^*$} & \underline{91.39$^*$} & 86.83 & 91.82 & \underline{\textbf{88.78}$^*$} & \underline{91.82$^*$} \\
            \fontsize{8pt}{5}{\textbf{MIARN3}} & 60.13 & 71.68 & 61.65 & 71.68 & 84.93 & 91.74 & 87.92 & 91.74 & 85.84 & 92.15 & 88.48 & 92.15\\
            \fontsize{8pt}{5}{\textbf{BTM3}} & 63.23 & 71.63 & 61.98 & 71.63 & 84.35 & 91.76 & 87.88 & 91.76 & 85.42 & 92.18 & 88.46 & 92.18\\
            \fontsize{8pt}{5}{\textbf{BERT3}} & 51.97 & 72.09 & 60.40 & 72.09 & 84.32 & \textbf{91.82} & 87.91 & \textbf{91.82} & 84.98 & \textbf{92.19} & 88.44 & \textbf{92.19} \\
            \fontsize{8pt}{5}{\textbf{GBP}} & 57.39 & 57.38 & 57.38 & 57.38 & 84.83 & 85.22 & 85.02 & 85.22 & 85.44 & 85.76 & 85.60 & 85.76 \\
            \fontsize{8pt}{5}{\textbf{HP}} & 81.05 & 76.82 & 78.20 & 76.82 & 92.20 & 79.54 & 82.65 & 79.54 & 93.01 & 92.90 & 92.89 & 92.89 \\
            \hline
        \end{tabular}
    }
\end{table*}
In this section, we present and discuss the experimental results of the research questions introduced in Section \ref{Experiment and Analysis}.

\begin{table*}[ht]
    \caption{\label{table: bi-task baseline} Baseline Results of Bi-Task.}
    \centering
    \resizebox{\textwidth}{!}{
        %        \begin{tabular}{p{1.5cm}p{0.8cm}p{0.8cm}p{0.8cm}p{0.8cm}p{0.8cm}p{0.8cm}p{0.8cm}p{0.8cm}p{0.8cm}p{0.8cm}p{0.8cm}p{0.8cm}}
        \begin{tabular}{ccccccccccccc}
            \hline
            \multirow{2}{*}{\textbf{Model}} & \multicolumn{4}{c}{Subtext Task} & \multicolumn{4}{c}{Metaphor Task} & \multicolumn{4}{c}{Sarcasm Task}  \\
            \cline{2-13}
            & \textbf{p(\%)}  & \textbf{r(\%)} & \textbf{F$_1$(\%)} & \textbf{acc(\%)} & \textbf{p$_m$(\%)} &\textbf{r$_m$(\%)} & \textbf{F$_{1_m}$(\%)} & \textbf{acc$_m$(\%)}&\textbf{p$_s$(\%)}  & \textbf{r$_s$(\%)} & \textbf{F$_{1_s}$(\%)} & \textbf{acc$_s$(\%)}  \\
            \hline
            \fontsize{8pt}{5}{\textbf{SASICMSS}} & 63.49 & 70.57 & 65.05 & 70.57 & - & - & - & - & 87.20 & 91.79 & \textbf{88.78} & 91.79 \\
            \fontsize{8pt}{5}{\textbf{MIARNSS}} & 61.94 & 71.61 & 61.94 & 71.61 & - & - & - & - & 85.53 & 92.14 & 88.47 & 92.14 \\
            \fontsize{8pt}{5}{\textbf{BTMSS}} & 60.66 & 71.51 & 62.02 & 71.51 & - & - & - & - & 84.98 & \textbf{92.19} & 88.44 & \textbf{92.19} \\
            \fontsize{8pt}{5}{\textbf{BERTSS}} & 51.97 & 72.09 & 60.40 & 72.09 & - & - & - & - & 84.98 & \textbf{92.19} & 88.44 & \textbf{92.19} \\
            \fontsize{8pt}{5}{\textbf{SASICMSM}} & 64.08 & 71.34 & 64.18 & 71.34 & 85.36 & 91.57 & 88.03 & 91.57 & - & - & - & - \\
            \fontsize{8pt}{5}{\textbf{MIARNSM}} & 60.37 & 71.65 & 61.72 & 71.65 & 84.67 & 91.77 & 87.90 & 91.77 & - & - & - & - \\
            \fontsize{8pt}{5}{\textbf{BTMSM}} & 60.57 & \textbf{72.11} & 61.27 & \textbf{72.11} & 84.32 & \textbf{91.82} & 87.91 & \textbf{91.82} & - & - & - & - \\
            \fontsize{8pt}{5}{\textbf{BERTSM}} & 51.97 & 72.09 & 60.40 & 72.09 & 84.32 & \textbf{91.82} & 87.91 & \textbf{91.82} & - & - & - & - \\

            \hline
        \end{tabular}
    }
\end{table*}
\begin{table*}[ht]
    \caption{\label{table: uni-task baseline} Baseline Results of Uni-Task.}
    \centering
    \resizebox{\textwidth}{!}{
        %        \begin{tabular}{p{1.5cm}p{0.8cm}p{0.8cm}p{0.8cm}p{0.8cm}p{0.8cm}p{0.8cm}p{0.8cm}p{0.8cm}p{0.8cm}p{0.8cm}p{0.8cm}p{0.8cm}}
        \begin{tabular}{ccccccccccccc}
            \hline
            \multirow{2}{*}{\textbf{Model}} & \multicolumn{4}{c}{Subtext Task} & \multicolumn{4}{c}{Metaphor Task} & \multicolumn{4}{c}{Sarcasm Task}  \\
            \cline{2-13}
            & \textbf{p(\%)}  & \textbf{r(\%)} & \textbf{F$_1$(\%)} & \textbf{acc(\%)} & \textbf{p$_m$(\%)} &\textbf{r$_m$(\%)} & \textbf{F$_{1_m}$(\%)} & \textbf{acc$_m$(\%)}&\textbf{p$_s$(\%)}  & \textbf{r$_s$(\%)} & \textbf{F$_{1_s}$(\%)} & \textbf{acc$_s$(\%)}  \\
            \hline

            \fontsize{8pt}{5}{\textbf{SASICMSubt}} & \textbf{63.58} & 71.19 & \textbf{63.37} & 71.19 & - & - & - & - & - & - & - & - \\
            \fontsize{8pt}{5}{\textbf{MIARNSubt}} & 60.70 & 71.67 & 61.13 & 71.67 & - & - & - & - & - & - & - & - \\
            \fontsize{8pt}{5}{\textbf{BTMSubt}} & 59.04 & 71.96 & 61.39 & 71.96 & - & - & - & - & - & - & - & - \\
            \fontsize{8pt}{5}{\textbf{BERT+FF}} & 51.98 & \textbf{72.10} & 60.41 & \textbf{72.10} & - & - & - & - & - & - & - & -\\
            \fontsize{8pt}{5}{\textbf{SVM}} & 60.67 & 72.00 & 60.60 & 72.00 & - & - & - & - & - & - & - & -\\
            \fontsize{8pt}{5}{\textbf{LR}} & 55.93 & 72.05 & 60.44 & 72.05 & - & - & - & - & - & - & - & -\\
            \fontsize{8pt}{5}{\textbf{MEC}} & 51.98 & \textbf{72.10} & 60.40 & \textbf{72.10} & - & - & - & - & - & - & - & -\\
            \fontsize{8pt}{5}{\textbf{NB}} & 61.14 & 11.50 & 13.35 & 11.50 & - & - & - & - & - & - & - & -\\
            \fontsize{8pt}{5}{\textbf{DT}} & 62.19 & 66.62 & 63.09 & 66.62 & - & - & - & - & - & - & - & -\\
            % \fontsize{8pt}{5}{\textbf{SASICMSarc}} & - & - & - & - & - & - & - & - & 87.29 & 90.40 & 88.59 & 90.40 \\
            % \fontsize{8pt}{5}{\textbf{MIARNSarc}} & - & - & - & - & - & - & - & - & 85.36 & 92.16 & 88.47 & 92.16 \\
            % \fontsize{8pt}{5}{\textbf{BTMSarc}} & - & - & - & - & - & - & - & - & 85.31 & 92.17 & 88.52 & 92.17 \\
            % \fontsize{8pt}{5}{\textbf{SASICMMeta}}& - & - & - & - & 86.71 & 90.37 & 88.01 & 90.37 & - & - & - & - \\
            % \fontsize{8pt}{5}{\textbf{MIARNMeta}} & - & - & - & - & 84.79 & 91.77 & 87.94 & 91.77 & - & - & - & - \\
            % \fontsize{8pt}{5}{\textbf{BTMMeta}} & - & - & - & - & 84.32 & 91.82 & 87.91 & \textbf{91.82} & - & - & - & - \\
            % \fontsize{8pt}{5}{\textbf{GBP}} & 57.39 & 57.38 & 57.38 & 57.38 & 84.83 & 85.22 & 85.02 & 85.22 & 85.44 & 85.76 & 85.60 & 85.76 \\
            % \fontsize{8pt}{5}{\textbf{HP}} & 81.05 & 76.82 & 78.20 & 76.82 & 92.20 & 79.54 & 82.65 & 79.54 & 93.01 & 92.90 & 92.89 & 92.89 \\
            %            \fontsize{8pt}{5}{\textbf{NaiveBayes}} & 53.62 & 58.41 & 48.48 & 58.41 & 85.98 & 1.74 & 0.5 & 1.74 & 68.07 & 73.92 & 70.70 & 73.92  \\
            \hline
        \end{tabular}
    }
\end{table*}
\subsubsection{Comparison with Baselines}
We compare four traditional baselines and three deep neural network baselines with SASICM. We extend the comparing model to tri-task structure by simply adding linear projection for each additional task as we did in SASICM. The results of tri-task model are shown in Table \ref{table: triple-task baseline}. Considering that BTM is a bi-task model, we decrease SASICM into bi-task framework by simply cutting an additional branch of corresponding task. In adddition, we also extend the uni-task model into bi-task structure for comparison. The results of bi-task model are shown in Table \ref{table: bi-task baseline}. Naturally, we compare all of the models under the uni-task settings. The results of uni-task are shown in Table \ref{table: uni-task baseline}. \par
Compared with single-task models, multi-task models have better performance in most of evaluation metrics, especially in the $F_1$ score. In particular, SASICM improves the performance in multi-tasks significantly more than other models.
As shown in Figure \ref{graph:classes number}, our data is extremely unbalanced, so we pay more attention to the $F_1$ score, especially when the accuracy rate is close to equal. $\mathrm{SASICM}_{g}$ gets the $F_1$ score of $64.37$, the third highest score in experiments. $\mathrm{SASICM}_{BERT}$ gets the $F_1$ score of $65.12$, the highest score in our experiments. We take SASICM based model as our baseline based on the following reasons: 1. SASICM$_g$ and  SASICM$_{BERT}$ achieve the third and first scores of the $F_1$ respectively under the premise of ensuring the accuracy; 2. it convergences faster than its variants, especially when comparing to the uni-task variant; 3. the number of parameters is 4\% less than MIARN3 whose parameters are more than 1.2 millions, and it is only one third of BTM3 whose parameters are more than 3 millions and it is only 10\% of BERT3 whose parameters are more than 1 billion.
\subsubsection{Comparison of Representation}
\begin{figure*}[!t]
    \centering
    \begin{minipage}[t]{\textwidth}
        \resizebox{\textwidth}{!}{
            \subfigure[Representation learnt by BERT ]{
                \includegraphics[width=0.5\textwidth]{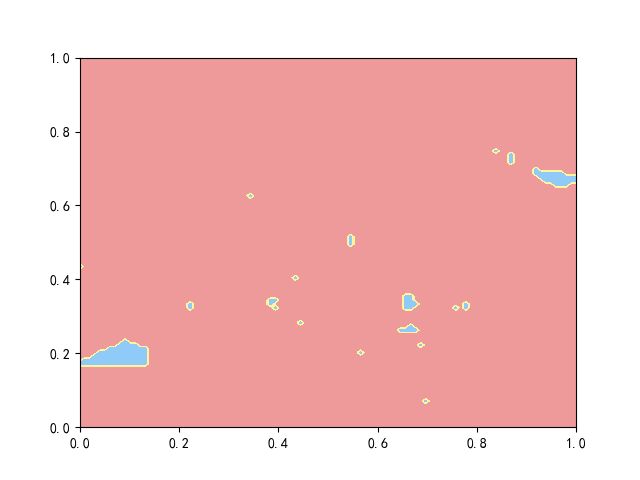}
                \label{tsne-BERT}
            } ~~~
            \subfigure[Representation learnt by BTM]{
                \includegraphics[width=0.5\textwidth]{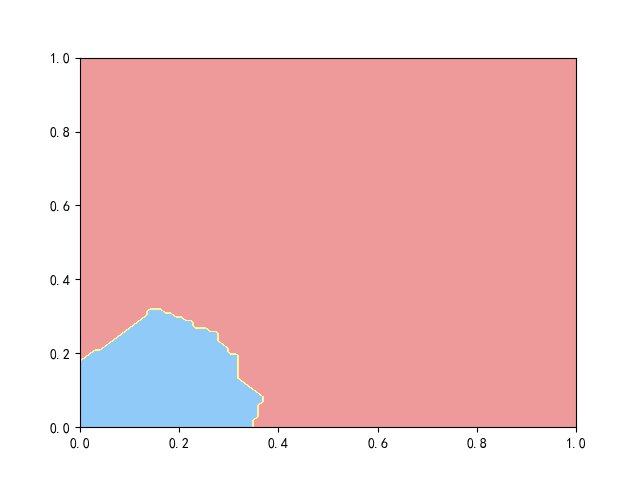}
                \label{tsne-BTM}
            }
        }
        \resizebox{\textwidth}{!}{
            \subfigure[Representation learnt by MIARN] {
                \includegraphics[width=0.5\textwidth]{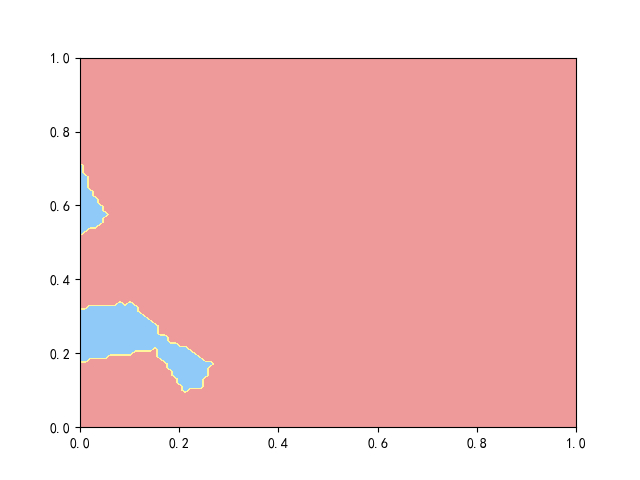}
                \label{tsne-MIARN}
            } ~~~
            \subfigure[Representation using BOW] {
                \includegraphics[width=0.5\textwidth]{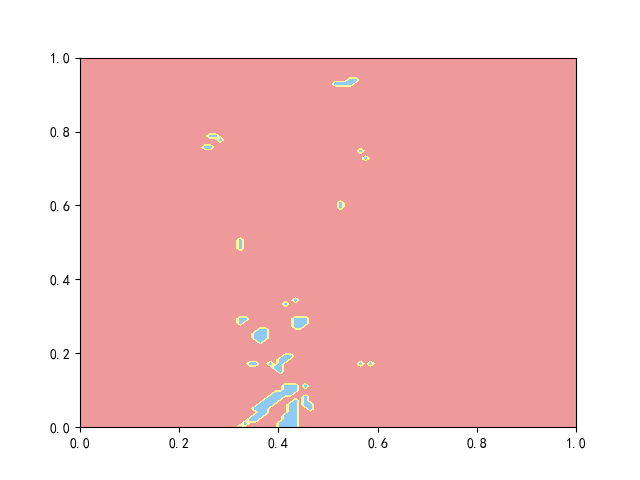}
                \label{tsne-BOW}
            }
        }
        \resizebox{\textwidth}{!}{
            \subfigure[Representation learnt by SASICM$_g$]{
                \includegraphics[width=0.5\textwidth]{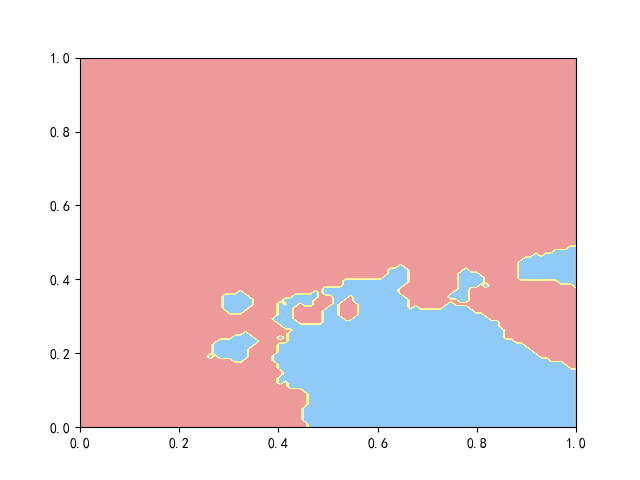}
                \label{tsne-SASICMg}
            } ~~~
            \subfigure[Representation learnt by SASICM$_bert$]{
                \includegraphics[width=0.5\textwidth]{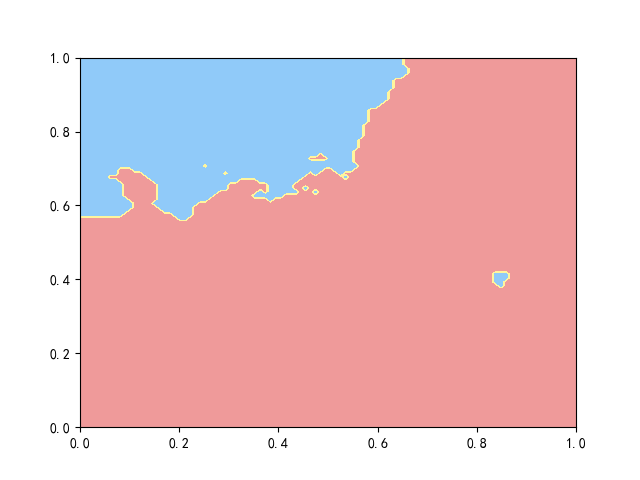}
                \label{tsne-SASICMbert}
            }
        }
    \end{minipage}
    \caption{Representations. The representations are learnt by different models, which is the outputs of penultimate layer in every model. We use t-SNE to reduce the dimension of representations, and we plot the results of them.}
    \label{Representations-TSNE}
\end{figure*}

This section, we show the representations learnt by different models in Figure \ref{Representations-TSNE}. The representation is the output of penultimate layer in every model. We reduce the dimension of representations by t-SNE \cite{JMLR:v9:vandermaaten08a}, which is a dimensionality reduction technique, then we use K-Means to learn the boundary of non-subtext and subtext. We show the boundary of subtext and non-subtext after dimension reduction in Figure \ref{Representations-TSNE}, where the pink region is a place without subtext, and the blue region is a place with subtext. Intuitively, if a representation can make the data points that belong to the same class get closer after clustering, it is good. Moreover, if the class has multiple patterns, there should be multiple centers after clustering. Therefore, we pay attention to the spatial size of the blue area and the distribution concentration of the blue area. The larger the blue area is, the better the representation is. The more of the number of the blue area, the better the representation is. It is meaningless to pay attention to the distribution of blue region, because different models and embedding methods will lead to different feature subspaces.

 As shown in Figure \ref{tsne-SASICMg} and \ref{tsne-SASICMbert}, SASICM$_g$ and SASICM$_{BERT}$ encode subtext into a larger space, and SASICM$_g$ learns more patterns than other models. BERT-based model cannot learn a good representation in subtext recognition tasks, as shown in Figure \ref{tsne-BERT}, which is due to the small spatial range of sub-text and its discrete spatial distribution. The method of Bag-Of-Word (BOW) has similar results like BERT as shown in Figure BOW. MIARN and BTM have learnt better representations than BERT and BOW bacause the spatial size of the blue region shown in Figure \ref{tsne-BTM} and \ref{tsne-MIARN} are much larger than that of BERT and BOW.

\subsubsection{Uni-Task Model Result}
Because MIARN is designed for uni-task and our subject is to recognize subtext, we execute \textbf{SASICMSubt}, \textbf{MIARNSubt}, \textbf{BTMSubt} and \textbf{BERT+FF} under uni-task setting. The results are shown in Table \ref{table: uni-task baseline}. The results show that SASICM is better than MIARN, BTM and BERT based model. In addition, our model are much better than the traditional classifiers, including Support Vector Machine (SVM), Logistic Regression (LR), Max Entropy Classifier (MEC), Naive Bayes classifier (NB) and Decision Tree (DT). The results show that most of the traditional classifiers can obtain the comparable performance as BERT, except NB, which means that using BOW as the input feature of NB is not suitable. SASICM needs about 5 hours to converge, the time of which is less than that of MIARN and BTM which need about 8 hours to converge and the time of which is much less than that of BERT that need about 24 hours to converge. Traditional classifiers are trained faster than SASICM. However, the ability of generalization of the traditional classifiers (NB, SVM, LR, MEC and DT) is weaker than SASICM. The main shortcomings of SVM, LR, MEC are that they have low precision scores. The $F_1$ score of DT are similar to SASICMSubt, however, the recall score of DT is too low.
\subsubsection{Multi-Task Model Result}
% 1. 更快，提供了更多的监督信息，加速收敛过程。比单任务的收敛速度要快一倍以上。
% 2. 可以提高f1值，减缓偏类问题。
From Table \ref{table: triple-task baseline} to Table \ref{table: uni-task baseline}, we observe that SASICM is improved more than BTM, MIARN and BERT under the same method of task feature confusion. The binary-task of subtext and sarcasm perform better than the binary-task of subtext and metaphor. The results of SASICM$_g$ and SASICM$_{BERT}$ show that triple-tasks framework is the trade-off between two binary-tasks: \textit{subtext + metaphor} and \textit{subtext + sarcasm}, and its performances of $F_1$ score and accuracy score are somewhere in between. Moreover, the triple-task model provides more supervision information, which can speed up the convergence. In addition, the increasing of $F_1$ score and no much decreasing of accuracy score reflect that triple-tasks can also alleviate overfitting problem by leveraging the metaphorical information and sarcasm information. 

\subsubsection{Different Embedding Methods}
We use different embedding methods for SASICM, which is marked as SASICM$_{g}$ and SASICM$_{BERT}$, respectively. The $F_1$ score of SASICM$_{BERT}$ is higher than that of SASICM$_g$, but the accuracy score of it is lower than that of SASICM$_g$. Both embedding approaches achieved similar results for the other two tasks. From Table \ref{table: triple-task baseline}, it can be observed that BERT can improve the precision score \textbf{p} and GloVe can improve the recall score \textbf{r}. This phenomenon shows that BERT, as a pre-training model, can find subtext better than GloVe, but the extracted subtext features are not as accurate as GloVe, and more noises are fitted than GloVe.\\
% 对比有RNN和没有RNN的结果
% The results are shown in Table \ref{table: baseline}. *SS and *SM are the corresponding models of subtext-sarcasm-task and subtext-metaphor-task, respectively. SASICMSt means that we feed the data into Bi-GRU before strengthen attention. SASICML means we use LSTM instead of GRU unit. SASICMSG means we use unidirectional GRU instead of bidirectional GRU. SASICMSA uses self-attention instead of strengthening attention. SASICMWC only uses cross entropy as loss of SASICM. *3 are tri-tasks frameworks. MIARN3 only linearly projects the output of last hidden layer, and obtains the features corresponding to the task before classification.  Metaphor and irony use stacked tensor matrix to generate interactive feature with subtexts, and then connect them with subtext features in series. *Subt are models executed on subtext task independently, *Sarc are models executed on sarcasm task, and *Meta are models executed on metaphor task. *$_m$ and $*_s$ are the corresponding results of metaphor detection and sarcasm detection. \par

\section{Other Ablation Study}
In this section, we explore the following problems:\\
\textbf{Why Self-Attention is not used:} Self-attention is widely used in many NLP models, such as Transformer \citep{DBLP:conf/nips/VaswaniSPUJGKP17}. In most cases, self-attention works very well when it comes to extracting features within a sentence. To this end, we conduct a related experiment, which is called SASICMSA. \\
\textbf{Why LSTM is not used:}  LSTM is the most widely sequential model. And it is used in MIARN as well. To this end, we conduct a corresponding experiment, which is called SASICML. \\
\textbf{Why constraint is needed:} Strengthen-Attention was first proposed in this paper. It is a question worth exploring whether or not it will have an effect without any constraints, and how it will have an effect. Therefore, we conduct a corresponding experiment called SASICMWC.\par
In addition, we explore some small questions, such as why a bidirectinal GRU is needed and why the internal feature extraction layer is not placed behind the GRU layer but simultaneously. The corresponding models are called SASICMSG and SASICMSt respectively. All the results of ablation study are shown in Table \ref{table: ablation}.
\begin{table*}[ht]
    
    \caption{\label{table: ablation} Ablation Model Results. The result of f1 score (F1) and accuracy (acc), where ``p'' is precision and ``r'' is recall. *$_m$ and *$_s$ are the results of metaphor and sarcasm respectively.}
    \centering
    \resizebox{\textwidth}{!}{
        %        \begin{tabular}{p{1.5cm}p{0.8cm}p{0.8cm}p{0.8cm}p{0.8cm}p{0.8cm}p{0.8cm}p{0.8cm}p{0.8cm}p{0.8cm}p{0.8cm}p{0.8cm}p{0.8cm}}
        \begin{tabular}{ccccccccccccc}
            \hline
            \multirow{2}{*}{\textbf{Model}} & \multicolumn{4}{c}{Subtext Task} & \multicolumn{4}{c}{Metaphor Task} & \multicolumn{4}{c}{Sarcasm Task}  \\
            \cline{2-13}
            & \textbf{p(\%)}  & \textbf{r(\%)} & \textbf{F$_1$(\%)} & \textbf{acc(\%)} & \textbf{p$_m$(\%)} &\textbf{r$_m$(\%)} & \textbf{F$_{1_m}$(\%)} & \textbf{acc$_m$(\%)}&\textbf{p$_s$(\%)}  & \textbf{r$_s$(\%)} & \textbf{F$_{1_s}$(\%)} & \textbf{acc$_s$(\%)}  \\
            \hline
            \fontsize{8pt}{5}{\textbf{SASICMSt}} & 63.21 & 71.38 & 63.60 & 71.38 & 85.10 & 91.61 & 88.01 & 91.61 & 86.81 & 92.02 & 88.63 & 92.02 \\
            \fontsize{8pt}{5}{\textbf{SASICML}} & 64.58 & 71.33 & 64.35 & 71.33 & 85.92 & 91.69 & \textbf{88.19} & 91.69 & 86.10 & 91.57 & 88.54 & 91.57 \\
            \fontsize{8pt}{5}{\textbf{SASICMWC}} & 62.79 & 71.67 & 62.32 & 71.67 & 84.61 & 91.78 & 87.92 & 91.78 & 87.19 & 92.19 & 88.53 & \textbf{92.19} \\
            \fontsize{8pt}{5}{\textbf{SASICMSG}} & 60.54 & 71.79 & 61.69 & 71.79 & 84.33 & 91.78 & 87.89 & 91.78 & 86.38 & 92.16 & 88.47 & 92.16 \\
            \fontsize{8pt}{5}{\textbf{SASICMSA}} & 64.11 & 71.67 & 63.38 & 71.67 & 85.19 & 91.63 & 87.96 & 91.63 & 86.87 & 91.90 & 88.61 & 91.90 \\
            \hline
        \end{tabular}
    }
\end{table*}

\subsection{Utility Of Strengthen Attention}
Results of SASICM in Table \ref{table: triple-task baseline} and SASICMSA in Table \ref{table: ablation} show that the $F_1$ score of strenghten attention with constraints is higher than that of self-attention. Take  ``黄诗扶最近是霸占了我的听觉！(Recently, Shifu Huang occupied my hearing!)'' for example, the maps of attention matrix are shown in Figure~\ref{AttentionValueOfSASICM} and \ref{AttentionValueOfSASICMSA}, where Figure~\ref{AttentionValueOfSASICM} denotes attention of SASICM and  Figure~\ref{AttentionValueOfSASICMSA} denotes attention of SASICMSA. Self-attention fails to pay attention to the word distinguishably in such example, while strengthen attention with restrictive conditions can support giving more distinguishable attention for other words and focus on important keywords more accurately.
\subsection{Why LSTM as Sequential Feature Extractor is not used}
The result of SASICM with LSTM (SASICML) is shown in Table \ref{table: ablation}. The result of SASICM with GRU (SASICM$_g$ and SASICM$_{BERT}$) is shown in Table \ref{table: triple-task baseline}. The $F_1$ score of SASICML is slightly lower than that of SASICM$_g$ and SASICM$_{BERT}$, and the accuracy score is slightly improved. It shows that LSTM and GRU are basically the same in terms of model effect. However, the training speed of LSTM is much slower than that of GRU. In our experiment, the training time of LSTM is nearly two hours slower than that of GRU, which is nearly 40\% more time. The reason for this is that the GRU structure is much simpler than the LSTM structure. Following Occam's razor principle, we chose the simpler structure as our sequential model. Moreover, we tried other more complicated structure for subtext, but it does not help to improve subtext recognition. Therefore, we restore current SASICM as our baseline.

\subsection{Utility Of Getting Attention Directly From Embedding}
From Table \ref{table: ablation} and Table \ref{table: triple-task baseline}, it can be found that the $F_1$ score of SASICM is better than that of SASICMSt and their accuracy score are similar, which proves that in our model, it is better to get attention directly from embedding layer than from RNNs. We sample 10k words from vocabulary, then we calculate the mean cosine similarity for SASICM and SASICMSt of the 10k words. The mean cosine similarity of SASICM is 0.1112. The mean cosine similarity of SASICMSt is 0.5211. The representation dimension in SASICMSt is twice as large as that in SASICM, but the mean cosine similarity of SASICMSt is about five times as large as that in SASICM. It proves that the representations after Bi-RNNs are similar, and it is better to get attention directly from embedding layer than from Bi-RNNs.
% \subsection{Utility Of Bi-GRU}
%     In this section, we analyze the effects of Bi-directional GRU (Bi-GRU). 
%     Table \ref{table: result} shows that bi-GRU performs much better than Uni-GRU on $F_1$ and precision, which means Bi-GRU can help to alleviate the problem of preferring to classifying the samples into the classes with the majority amount of instances. One of the possible reasons for this phenomena can be that bi-GRU not only model the forward sequence features, but also model the backward sequence features. Bi-direction information can be combined to find the keywords more accurately which appears many times in a text. Moreover, these keywords are likely to be the keywords of subtext, metaphor or sarcasm.

\begin{figure*}[!t]
    \centering
    \begin{minipage}[t]{0.9\textwidth}
        \resizebox{\textwidth}{!}{
            \subfigure[Attention of SASICM]{
                \includegraphics[width=0.5\textwidth]{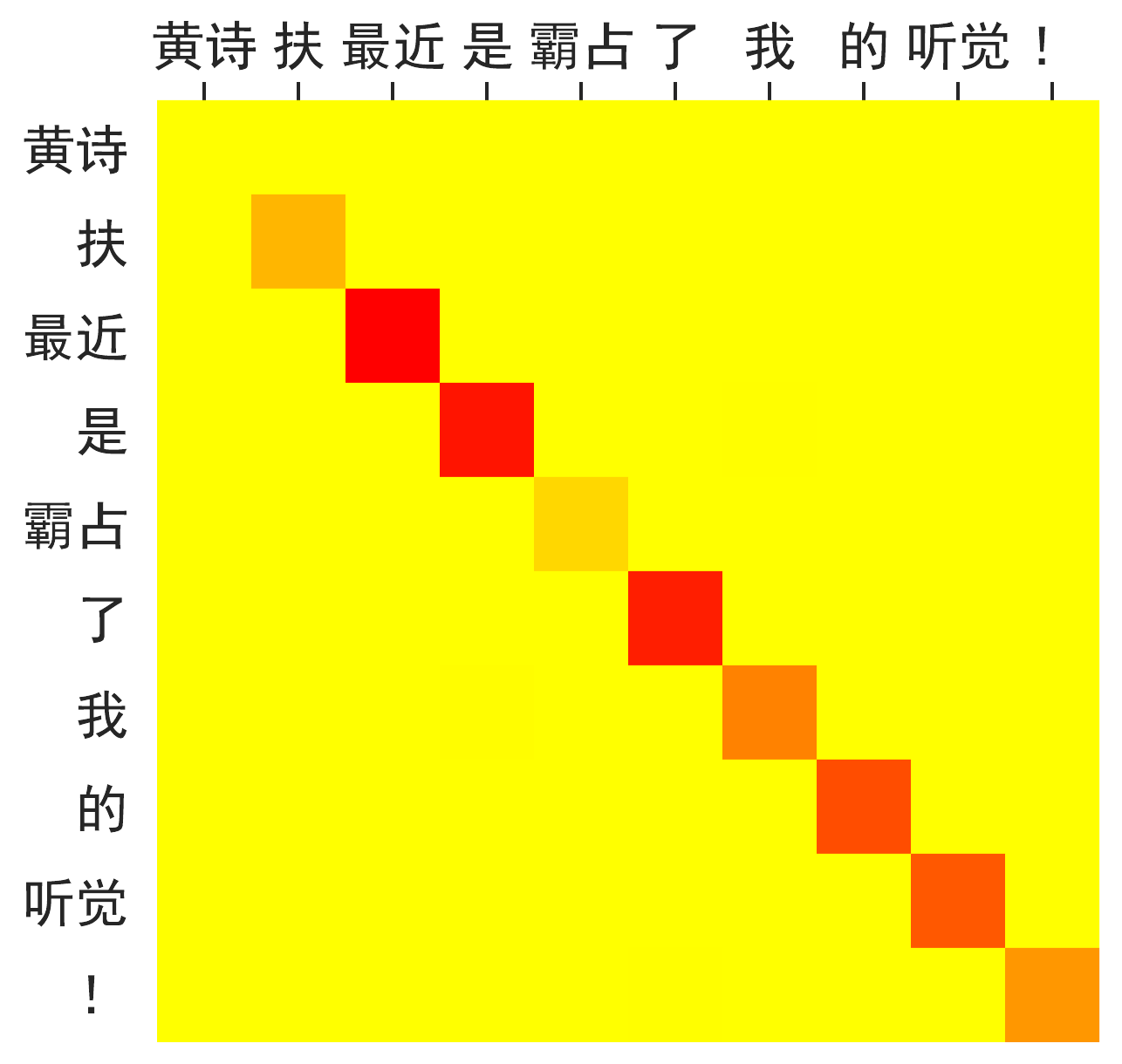}
                \label{AttentionValueOfSASICM}
            }~~~~
            \subfigure[Attention of SASICMSA]{
                \includegraphics[width=0.5\textwidth]{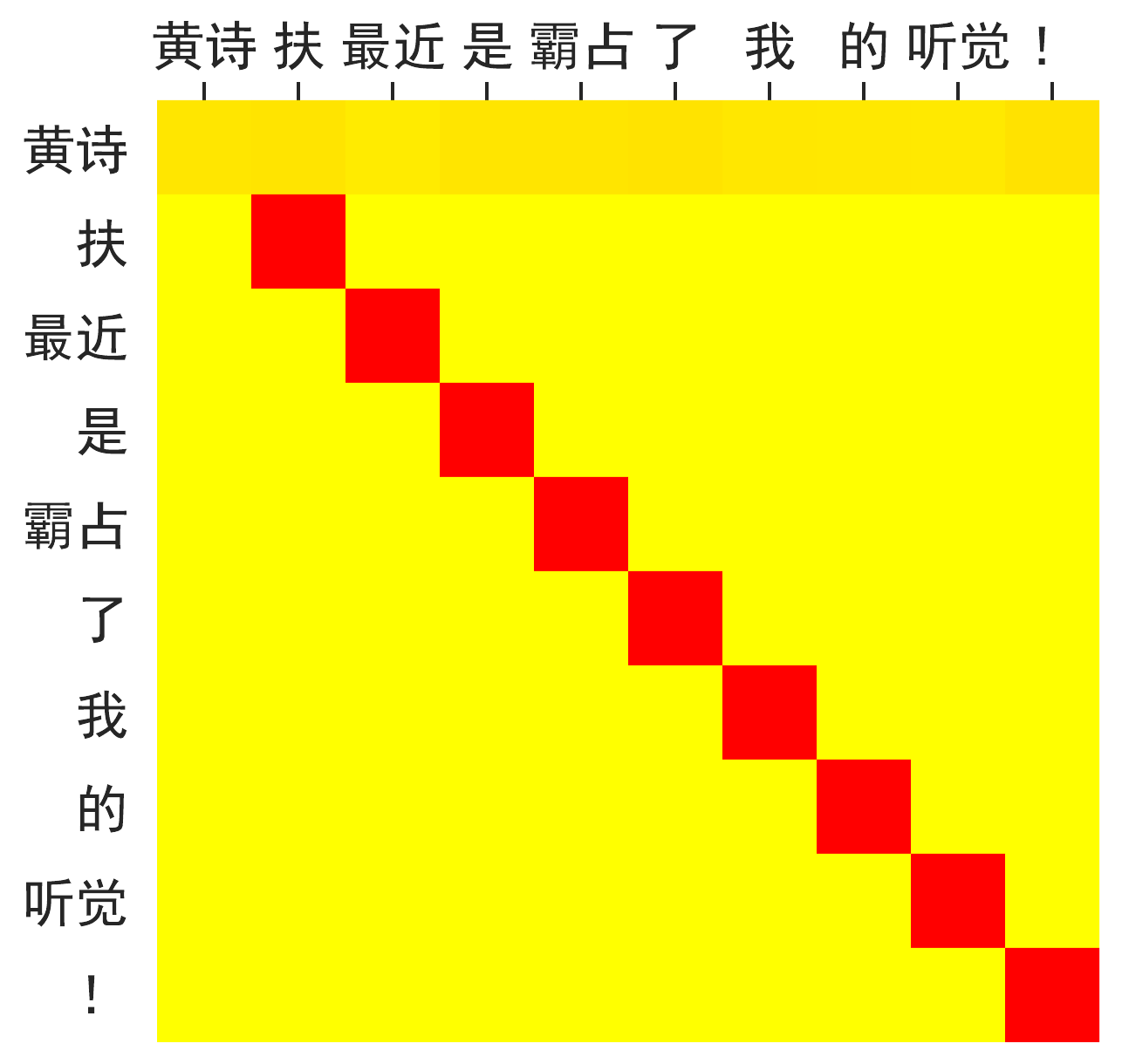}
                \label{AttentionValueOfSASICMSA}
            }
        }
    \end{minipage}
    \caption{The maps of attention matrix for ``黄诗扶最近是霸占了我的听觉！'', and the larger the attention, the darker the red.}
    \label{attention-example}
\end{figure*}

\subsection{Utility Of Constraints}
Results of SASICM and SASICMWC show that $F_1$ score of strengthen attention with constraints is much higher than that of without constraints. From the experiment, the difference between SASICM with constraints (SASICM) and SASICM without constraints (SASICMWC) is that the constraints can make the parameters of strengthen attention ($t_t$ and $c$), converge to the desired interval. The SASICMWC is highly random, resulting that the value of $t_t$ can be small and the value of $c$ will be close to 0, which means that nearly all words are important. Therefore, SASICM without constraints can not achieve the goal that maximizing the difference between important words and unimportant words, and expanding the difference of attentions.
% 对比用Multi-Task和不用Multi-Task的结果
% \subsection{Utility Of Multi-Task}

\section{Conclusion}
Subtext is a deep semantic meaning that is more difficult to get than sarcasm and metaphor. We collected the data from the popular social media, then we constructed a Chinese dataset for subtext recognition problem. To deal with subtext recognition, We built SASICM, our proposed method, which obtains the $F_1$ score and accuracy of 64.37\% and 71.11\% respectively. 
\section{Future Work}
Results in Table \ref{table: triple-task baseline} show that there is still room for improvement in subtext recognition. From the case analysis, we can improve the study from the following aspects: First, reduce the number of incorrect word segmentations, as shown in Figure~\ref{attention-example}. The name of people is wrong segmented into two words ``黄诗/扶~(Huangshi/Fu)'' which actually should be ``黄诗扶~(Shifu Huang)''. This will degrade the performance of SASICM.
Second, the attention value of SASICM and SASICMSA have a common shortcoming that nearly all the words in text pay too much attention to themselves. Therefore, making each words pay more attention to other words may be a way to improve the performance. Last, we shall perform more fine-grained tasks, such as judging the type of subtext. In addition, limited by the existing data resource, our current work only performs experiments with informal text used in social media, and our goal is to find the obvious subtext in text. In the future, we will devote ourselves to solving these problems, and try to analyze subtext in the official text.
\section{Acknowledgements}
This work is supported in part by the National Science Foundation of China under Grant Nos. (61876076).
\end{CJK*}

\bibliography{mybibfile}
%\bibliography{anthology}

\end{document}